\documentclass[final]{cvpr}

\usepackage{times}
\usepackage{epsfig}
\usepackage{graphicx}
\usepackage{amsmath}
\usepackage{amssymb}

\usepackage{multirow}
\usepackage{booktabs}
\usepackage{verbatim}
\usepackage{subfig}
\usepackage{bm}

\def\vp{{\bm{p}}}

\def\vg{{\bm{g}}}

\usepackage{xcolor}
\definecolor{bittersweet}{rgb}{1.0, 0.44, 0.37}
\definecolor{mygreen}{rgb}{0.29, 0.7, 0.48}

\usepackage[font=small,labelfont=bf]{caption}  
\usepackage{makecell}
\usepackage{tabulary}
\definecolor{demphcolor}{RGB}{144,144,144}
\newcommand{\demph}[1]{\textcolor{demphcolor}{#1}}
\definecolor{mygray}{gray}{0.4}
\usepackage{pifont}
\newcommand{\cmark}{\color{mygray}\ding{51}}%
\newcommand{\xmark}{\color{mygray}\ding{55}}%
\newlength\savewidth\newcommand\shline{\noalign{\global\savewidth\arrayrulewidth
  \global\arrayrulewidth 1pt}\hline\noalign{\global\arrayrulewidth\savewidth}}

\newcommand{\tablestyle}[2]{\setlength{\tabcolsep}{#1}\renewcommand{\arraystretch}{#2}\centering\footnotesize}
\makeatletter\renewcommand\paragraph{\@startsection{paragraph}{4}{\z@}
  {.5em \@plus1ex \@minus.2ex}{-.5em}{\normalfont\normalsize\bfseries}}\makeatother
\def\x{$\times$} 
\def\ModelName{\textsc{ClipBERT}}

\newcolumntype{C}[1]{>{\centering\arraybackslash}p{#1}}
\newcolumntype{R}[1]{>{\raggedleft\arraybackslash}p{#1}}
\newcolumntype{L}[1]{>{\raggedright\arraybackslash}p{#1}}

\usepackage[pagebackref=true,breaklinks=true,colorlinks,bookmarks=false]{hyperref}

\begin{document}

\title{Less is More: \ModelName~for Video-and-Language Learning \\ via Sparse Sampling}

\author{Jie Lei\thanks{\,\, Equal contribution.}\,\,$^1$, Linjie Li\footnotemark[1]\,\,$^2$, Luowei Zhou$^2$, Zhe Gan$^2$, Tamara L. Berg$^1$, Mohit Bansal$^1$, Jingjing Liu$^2$\\ 
  $^1$UNC Chapel Hill \quad $^2$Microsoft Dynamics 365 AI Research \\
  \texttt{\small\{jielei, tlberg, mbansal\}@cs.unc.edu} \\
  \texttt{\small\{lindesy.li, luowei.zhou, zhe.gan, jingjl\}@microsoft.com}
  }

\maketitle

\begin{abstract}
The canonical approach to video-and-language learning (\eg, video question answering) dictates a neural model to learn from offline-extracted dense video features from vision models and text features from language models. 
These feature extractors are trained independently and usually on tasks different from the target domains, rendering these fixed features sub-optimal for downstream tasks.
Moreover, due to the high computational overload of dense video features, it is often difficult (or infeasible) to plug feature extractors directly into existing approaches for easy finetuning.
To provide a remedy to this dilemma, we propose a generic framework \ModelName~that enables affordable end-to-end learning for video-and-language tasks, by employing \textbf{sparse sampling}, where only a single or a few sparsely sampled short clips from a video are used at each training step.
Experiments on text-to-video retrieval and video question answering on six datasets demonstrate that \ModelName~outperforms (or is on par with) existing methods that exploit full-length videos, suggesting that end-to-end learning with just a few sparsely sampled clips is often more accurate than using densely extracted offline features from full-length videos, proving the proverbial \textbf{less-is-more} principle.
Videos in the datasets are from considerably different domains and lengths, ranging from 3-second generic-domain GIF videos to 180-second YouTube human activity videos, showing the generalization ability of our approach.
Comprehensive ablation studies and thorough analyses are provided to dissect what factors lead to this success.
Our code is publicly available.\footnote{\url{https://github.com/jayleicn/ClipBERT}}
\end{abstract}

\vspace{-3mm}
\section{Introduction}\label{sec:intro}

\begin{figure}[!t]
  \centering
  \includegraphics[width=\linewidth]{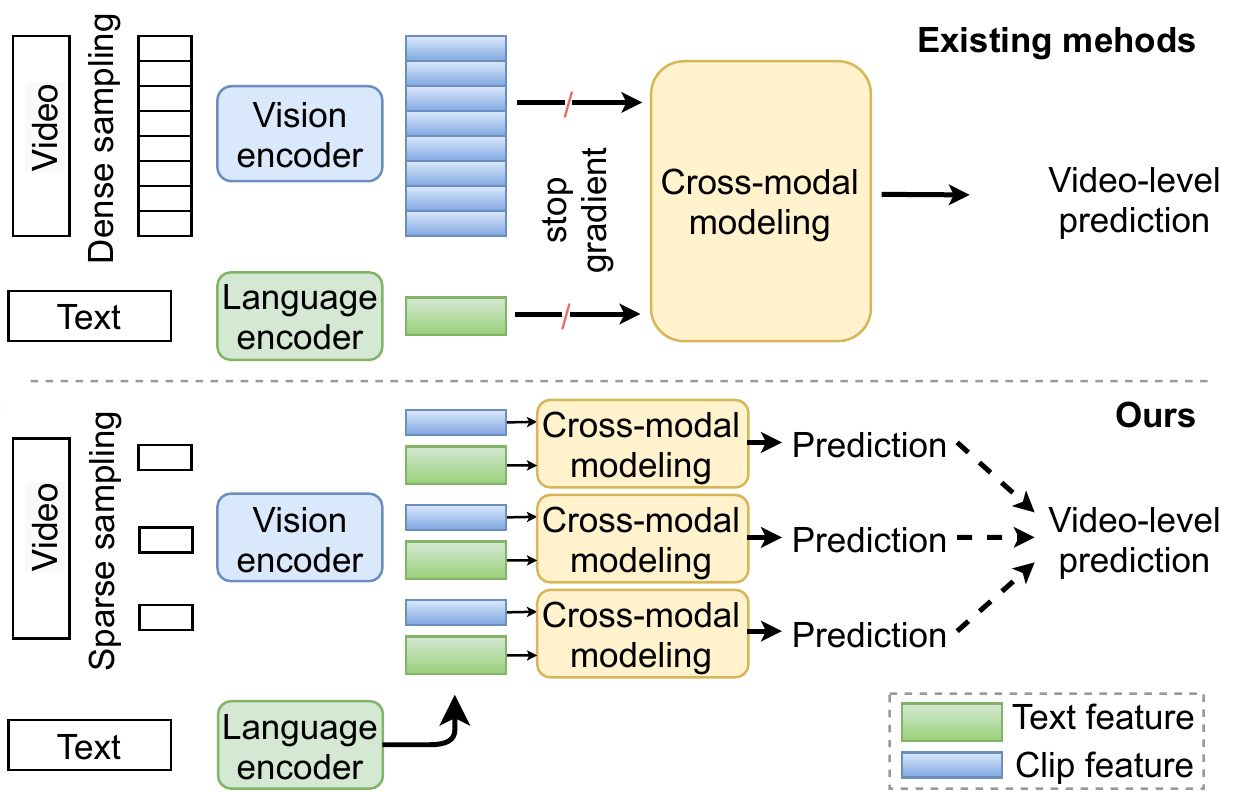}
  \caption{ 
Comparison between popular video-and-language learning paradigm (\textit{top}) and \ModelName~(\textit{bottom}).
In contrast to most existing methods that utilize offline (\textit{stop gradient}) extracted \textit{dense video features} and \textit{text features}, \ModelName~uses \textit{sparsely sampled clips} and \textit{raw text tokens} for \textit{end-to-end} modeling.
}
  \label{fig:ClipBERT_comparison}
  \vspace{-3mm}
\end{figure}

Humans communicate with each other in this interactive and dynamic visual world via languages, signs, and gestures. 
The ability to jointly understand both visual and textual clues is an essential ability for intelligent agents to interpret multimodal signals in the physical world.
A wide range of tasks based on real-life videos have been designed to test such ability, including text-to-video retrieval~\cite{xu2016msr, krishna2017dense, rohrbach2015dataset}, video captioning~\cite{rohrbach2015dataset,xu2016msr,zhou2017towards}, video question answering~\cite{xu2017video, jang2017tgif, lei2018tvqa,lei2019tvqa+}, and video moment retrieval~\cite{anne2017localizing, gao2017tall, lei2020tvr}. 
The \emph{de facto} paradigm to tackle these cross-modal tasks is to first extract dense video features from pre-trained vision models~\cite{he2016deep,carreira2017quo} and text features from pre-trained language models~\cite{pennington2014glove,devlin2018bert}, then apply multimodal fusion to wrangle together these fixed representations in a shared embedding space (Figure~\ref{fig:ClipBERT_comparison} (\textit{top})).

Existing approaches~\cite{jang2017tgif,lei2018tvqa,zhu2020actbert,le2020hierarchical} following this paradigm have achieved strong success, yet suffer from two main drawbacks: $(i)$ \emph{Disconnection in tasks/domains}: offline feature extractors are often trained on tasks and domains different from the target task. For example, features learned for action recognition from human activity videos~\cite{kay2017kinetics} are incongruently applied to downstream video question answering on generic-domain GIF videos~\cite{jang2017tgif}. $(ii)$ \emph{Disconnection in multimodal features}: learned features from different modalities are independent of each other. For instance, action recognition models~\cite{tran2015learning, wang2016temporal, carreira2017quo} are typically trained from pure video data without textual input, yet are applied to video-and-language tasks.
End-to-end task-specific finetuning offers a way to mitigate these inherent disconnections.
However, extracting features from the full sequence of video frames, as in most existing work, casts excessive demand on memory and computation, rendering it difficult or even infeasible to directly plug feature extractors into a video+language learning framework for efficient end-to-end finetuning.

Motivated by this, we propose \ModelName, a generic and efficient framework for end-to-end video-and-language learning (Figure~\ref{fig:ClipBERT_comparison} (\textit{bottom})).
Two aspects distinguish \ModelName~from previous work.
First, in contrast to densely extracting video features (adopted by most existing methods), \ModelName~sparsely samples only one single or a few short clips from the full-length videos at each training step. The hypothesis is that visual features from sparse clips already capture key visual and semantic information in the video, as consecutive clips usually contain similar semantics from a continuous scene. Thus, a handful of clips are sufficient for training, instead of using the full video.
Then, predictions from multiple densely-sampled clips are aggregated to obtain the final video-level prediction during inference, which is less computational demanding.
This \emph{sparse-training}-then-\emph{dense-inference} strategy greatly reduces memory needs and computations, allowing economical end-to-end learning from raw video frame pixels and language tokens.

The second differentiating aspect concerns the initialization of model weights (\ie, transfer through pre-training). 
In recent literature, image-text pre-training (\eg, using COCO Captions~\cite{chen2015microsoft} or Visual Genome Captions~\cite{krishna2017visual}) has been applied to image-text tasks~\cite{tan2019lxmert,lu2019vilbert,chen2019uniter,su2019vl,huang2020pixel,li2020unicoder,zhou2020unified}, and video-text pre-training (\eg, using HowTo100M~\cite{miech2019howto100m}) to video-related tasks~\cite{sun2019videobert,zhu2020actbert,gabeur2020multi,li2020hero}. 
There has been no study to cross-examine the effect of image-text pre-training on video-text tasks. Intuitively, visual features learned through pre-training from large-scale image datasets should also help video understanding tasks that rely on visual clues in static video frames.
To investigate this, we use 2D architectures (\eg, ResNet-50~\cite{he2016deep}) instead of 3D features ~\cite{tran2015learning,carreira2017quo,qiu2017learning,Xie_2018_ECCV} as our visual backbone for video encoding, allowing us to harness the power of image-text pre-training for video-text understanding along with the advantages of low memory cost and runtime efficiency.
Empirically, we observe that the knowledge learned in image-text pre-training indeed helps video-text tasks; this simple strategy helps us achieve better or comparable performance to previous state of the art on text-to-video retrieval and video question answering tasks.

Our contributions are three-fold:
($i$) We propose \ModelName, a new end-to-end learning framework for video+language tasks. 
Experiments show that \ModelName~achieves superior (or on par) performance than existing methods across diverse video-text tasks, where the average video length ranges from a few seconds to three minutes.
($ii$) Our work suggests \textit{``less is more''}: the proposed end-to-end training strategy with a single or a few (\textit{less}) sparsely sampled clips is often \textit{more} accurate than traditional approaches that employ densely extracted video features.
($iii$)  We demonstrate that image-text pre-training benefits video-text tasks.
We also provide comprehensive ablation studies to reveal the key factors that lead to the success of \ModelName, in hope of inspiring more future work.

\section{Related Work}\label{sec:related_work}
\paragraph{Video and Language Understanding.}  
Popular video-and-language tasks include text-to-video retrieval~\cite{xu2016msr, krishna2017dense, rohrbach2015dataset}, video captioning~\cite{xu2016msr, zhou2017towards, krishna2017dense, rohrbach2015dataset, li2016tgif}, video question answering~\cite{xu2017video, jang2017tgif, lei2018tvqa}, and moment retrieval~\cite{anne2017localizing, gao2017tall, lei2020tvr}. 
Standard approaches~\cite{jang2017tgif, xu2017video, gao2018motion, zhang2018cross, lei2018tvqa, fan2019heterogeneous, le2020hierarchical,lei2020mart} leverage offline extracted video and text features from action recognition models~\cite{kay2017kinetics, wang2016temporal, carreira2017quo, Xie_2018_ECCV}, image recognition models~\cite{deng2009imagenet, he2016deep}, and language models~\cite{mikolov2013distributed, pennington2014glove, devlin2018bert, liu2019roberta}.
Aligned with the success of transformer-based~\cite{vaswani2017attention} language pre-training~\cite{devlin2018bert,liu2019roberta,yang2019xlnet,raffel2020exploring,lan2019albert,clark2020electra} and image-text pre-training~\cite{tan2019lxmert,lu2019vilbert,chen2019uniter,li2020unicoder,huang2020pixel,zhou2020unified,gan2020large,cho2021unifying},
video-text pre-training ~\cite{sun2019videobert,zhu2020actbert,gabeur2020multi,li2020hero,miech2020end,miech2019howto100m} has shown promising results on video-and-language tasks.
Beyond using fixed features and same-domain data (\ie, video-text pre-training only for video-text tasks), our work focuses on end-to-end training and applying image-text pre-training for video-text tasks.

\paragraph{Action Recognition.}
Modern video action recognition architectures are typically designed with deep 2D~\cite{simonyan2014very,szegedy2015going,he2016deep} or 3D~\cite{tran2015learning,carreira2017quo,Xie_2018_ECCV} convolutional networks. 
These backbones are often computation and memory heavy, making it extremely difficult to directly process videos of considerable length.
To ease this difficulty, instead of training on full-length videos, models are often trained with randomly sampled short clips from the videos~\cite{simonyan2014two,tran2015learning,qiu2017learning,Xie_2018_ECCV,wang2018non,feichtenhofer2019slowfast,feichtenhofer2020x3d,wang2016temporal}.
At inference time, predictions from multiple uniformly sampled clips are aggregated together as the final video-level prediction.
In relation to these works, we adopt a similar strategy to perform sparse training and dense inference to reduce overhead on video processing, but focus on video-and-language tasks with cross-modal modeling of video and language, in contrast to pure video modeling.

\section{\ModelName~with Sparse Sampling}
We propose \ModelName, a general framework that enables end-to-end learning on video and language data, by learning joint representations directly from video frame pixels and raw text tokens, instead of from offline-extracted single-modality features.
Figure~\ref{fig:ClipBERT_comparison} (\textit{bottom}) gives an overview of \ModelName~framework. 
It adopts a sparse sampling strategy using only a single or a few sampled clips at each training step, instead of full-length videos.
Each sampled clip is independently encoded with a vision backbone model, the visual features from which are then fed to a cross-modal module that extracts relations between the clip and its associated text representations.
Independent predictions from all the sampled clips are fused together (\eg, \textit{through mean-pooling}) to derive a consensus at the video level.
A task-specific loss is calculated based on this consensus to learn model parameters.
During inference, \ModelName~densely samples a sequence of clips from the video and aggregates their predictions as the final prediction.

Most existing work~\cite{jang2017tgif,lei2018tvqa,zhu2020actbert,le2020hierarchical} models offline-extracted dense video features and text features. 
Formally, we denote a video-text pair as $V$ (for video) and $S$ (for text sequence). The video $V$ is further denoted as a list of $N$ clips of equal duration $[c_1, c_2, ..., c_N]$. 
This standard paradigm can be formulated as:
\begin{align}\label{eq:existing_approach_modeling}
    p \,{=}\, \mathcal{H}([\mathcal{F}^{SG}_{v}(c_1), 
    \mathcal{F}^{SG}_{v}(c_2), 
    ..., \mathcal{F}^{SG}_{v}(c_N)],
    \mathcal{F}^{SG}_{l}(S)),
\end{align}
\noindent
where $\mathcal{F}^{SG}_{v}$ and $\mathcal{F}^{SG}_{l}$ are vision and language encoder, respectively. The superscript $SG$ denotes \textit{Stop Gradient}, meaning that gradients cannot be back-propagated through the two encoders. 
$\mathcal{H}$ is a cross-modal encoder and predictor, which models the relations between the encoded video/language inputs and makes predictions. 
$p$ is the video-level prediction. A task-specific loss function $\mathcal{L}^{task}$ is then applied to calculate the loss value $l^{task}$ based on this prediction and its corresponding ground-truth $q$:
\begin{align}\label{eq:existing_approach_loss}
    l^{task} = \mathcal{L}^{task}(p, q).
\end{align}

\paragraph{Sparse Sampling for Training.}
Instead of using the full-length video with $N$ clips, \ModelName~sparsely (and randomly) samples $N_{train}$ clips $\{c_{\tau_{i}}\}_{i=1}^{N_{train}}$ from $V$ for training. 
$N_{train}$ is typically much smaller than $N$. 
We model a sampled clip $c_{\tau_{i}}$ together with text input $S$ to produce a prediction $p_{\tau_{i}}$: 
\begin{align}\label{eq:clip_bert_modeling}
    p_{\tau_{i}} = \mathcal{H}(\mathcal{F}_{v}(c_{\tau_{i}}), \mathcal{F}_{l}(S)),
\end{align}

\noindent
where $\mathcal{F}_{v}$ and $\mathcal{F}_{l}$ are vision/language encoders. 
Different from Equation~\ref{eq:existing_approach_modeling} that uses offline vision/language encoders, \ModelName~is end-to-end trainable, allowing task-specific loss to further finetune the encoders, learning better representations. 
Independent predictions from all sampled clips are aggregated to derive a consensus. The loss value $l^{task}$ is calculated based on this video-level consensus:
\begin{align}\label{eq:clip_bert_loss}
    l^{task} = \mathcal{L}^{task}(\mathcal{G}(p_{\tau_{1}}, p_{\tau_{2}}, ..., p_{\tau_{N_{train}}}), q),
\end{align}

\noindent
where $\mathcal{G}$ is the prediction/score aggregation function (\eg, mean-pooling). 
When $N_{train}$ is larger than one, this formulation can be regarded as a form of multiple instance learning (MIL)~\cite{wu2015deep}.
At inference, we uniformly sample $N_{test}$ clips of the same duration as training clips, then aggregate predictions from all $N_{test}$ clips to form our final prediction. 

\ModelName's sparse training strategy can be interpreted as a type of data augmentation: different subsets of clips from the same video are used at different training steps, which improves the model's generalization ability.
In this sense, it is analogous to random cropping~\cite{simonyan2014very,he2016deep} commonly used in image classification tasks. 
It also takes inspiration from action recognition methods~\cite{simonyan2014two,tran2015learning,wang2016temporal,feichtenhofer2019slowfast}, where a video classifier is trained on sampled clips.

\begin{figure}[!t]
  \centering
  \includegraphics[width=\linewidth]{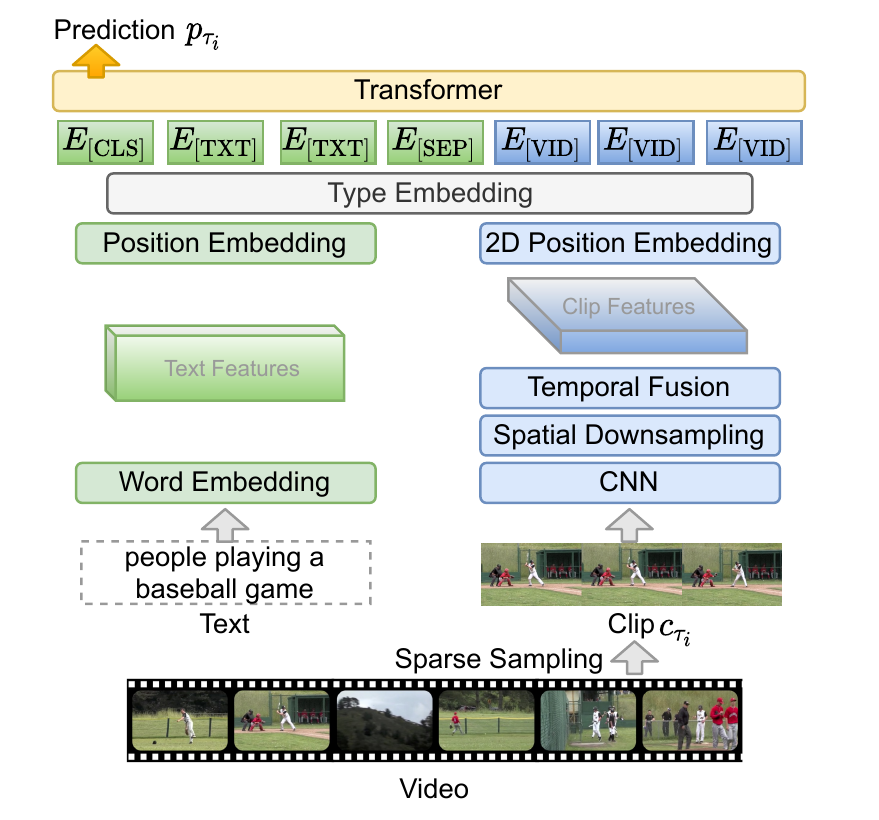}
  \vspace{-15pt}
  \caption{
Overview of \ModelName~architecture. For simplicity, we only show an illustration of producing prediction for a single sampled clip. When multiple clips are used, their predictions are fused together as the final prediction. 
}
  \label{fig:ClipBERT_instantiation}
  \vspace{-3mm}
\end{figure}

\paragraph{Model Architecture.}
Figure~\ref{fig:ClipBERT_instantiation} gives an overview of \ModelName~architecture. 
For the vision encoder $\mathcal{F}_{v}$, we use a 2D CNN architecture ResNet-50~\cite{he2016deep} instead of 3D architectures (such as C3D~\cite{tran2015learning} or I3D~\cite{carreira2017quo}), because 2D models typically consume less memory and run faster. 
Besides, 2D CNNs have proved to work reasonably well on video understanding tasks such as action recognition~\cite{wang2016temporal,qiu2017learning}.
Specifically, we take the first 5 Conv blocks of ResNet-50~\cite{he2016deep} and add an extra convolution layer to reduce its output feature depth, as well as a 2\x2 max-pooling layer for spatial down-sampling, following Pixel-BERT~\cite{huang2020pixel}. 
For each sampled clip, we uniformly sample $T$ frames and obtain $T$ feature maps. 
A temporal fusion layer $\mathcal{M}$ (\eg, mean-pooling) is applied to aggregate the frame-level feature maps into a single clip-level feature map. 
We then add a row-wise and a column-wise position embedding to each feature vector based on their 2D position. 
These embeddings are the same trainable position embeddings as in BERT~\cite{devlin2018bert}.
Collectively, these two position embeddings are indicative of 2D spatial locations of the features, which can be viewed as a \textit{2D position embedding}. 
The resulting feature map is flattened into an embedding sequence to represent the clip.

We use a trainable word embedding layer as our language encoder $\mathcal{F}_{l}$ to encode language tokens and add trainable position embeddings to encode positional information of the tokens.
Next, we add different type embeddings~\cite{devlin2018bert} to both clip and text embeddings to indicate their source type.
These two sequences are then concatenated as inputs to a 12-layer transformer~\cite{vaswani2017attention,devlin2018bert} for cross-modal fusion.
Special tokens \texttt{[CLS]} and \texttt{[SEP]} are added in this process following \cite{devlin2018bert}.
Given a downstream task, we add a task-specific prediction head with the last-layer \texttt{[CLS]} representation as input (\eg, using a two-layer MLP with softmax to produce scores for text-to-video retrieval).

\paragraph{Weight Initialization and Pre-training.}
We initialize the ResNet-50 layers with weights from grid-feat~\cite{jiang2020defense,ren2015faster}. It is trained on Visual Genome~\cite{krishna2017visual} for object detection and attribute classification, and produces effective grid features for image VQA tasks~\cite{antol2015vqa,gurari2018vizwiz}. 
Input frames are resized to have a maximum longer side of $L$ while keeping the aspect ratios, and the shorter side is zero-padded to be $L$ as well~\cite{nguyen2020revisiting}. 
We initialize the transformer and word embedding layers from BERT-base model~\cite{devlin2018bert}, pre-trained on BookCorpus~\cite{zhu2015aligning} and English Wikipedia.
These weights are trained separately for their individual single-modality tasks, thus simply combining them together in a single framework for downstream task training may result in sub-optimal performance. 
Although pre-training the whole model end-to-end with large-scale video-text datasets such as HowTo100M~\cite{miech2019howto100m} are appealing, we are restricted by the enormous computation cost.\footnote{\cite{miech2020end} reports that pre-training I3D~\cite{carreira2017quo} with offline extracted text features on HowTo100M requires $\sim$3 days with 64 Cloud TPUs v3. 
}
Luckily, as we use 2D CNN as our vision encoder, \ModelName~is able to directly take image-text pairs as inputs for training.
Thus, we leverage large-scale image-text datasets (COCO Captions~\cite{chen2015microsoft} and Visual Genome Captions~\cite{krishna2017visual}) to perform cross-modal pre-training~\cite{tan2019lxmert,lu2019vilbert,huang2020pixel}.
Specifically, we use masked language modeling~\cite{devlin2018bert} and image-text matching~\cite{tan2019lxmert,lu2019vilbert} objectives to optimize the model. 
By default, we finetune our model from this pre-trained weights for downstream video-text tasks. 
The impact of different weight initialization strategies is examined in Section~\ref{subsec:analysis_pretraining_and_e2e_training}.

\paragraph{Implementation Details.}
We perform image-text pre-training on COCO Captions~\cite{chen2015microsoft} and Visual Genome Captions~\cite{krishna2017visual}.
These two datasets contain a total of 5.6M training image-text pairs on 151K images.
This is the same data used in UNITER's~\cite{chen2019uniter} in-domain pre-training.
We use input image size $L{=}768$, and the resulting feature map from the vision encoder contains 144 pixels.
To improve generalization and reduce computation cost, during pre-training, we follow Pixel-BERT~\cite{huang2020pixel} to use pixel random sampling that samples 100 pixels from the encoded feature map as the input to the transformer layers.
Note that we only apply pixel random sampling for pre-training, and always use the full feature map for downstream tasks to avoid misalignment in training and inference~\cite{huang2020pixel}.
We use WordPiece embeddings~\cite{wu2016google} and keep at most 20 tokens from the caption. 
We then randomly mask 15\% of the tokens for masked language modeling. 
For each image-caption pair, with a probability of 0.5, we replace the ground-truth caption with a randomly sampled caption from another image to form a negative pair for image-text matching.
We use AadmW~\cite{loshchilov2017decoupled} to optimize end-to-end model training, with an initial learning rate of 5e-5, $\beta_1{=}0.9$, $\beta_2{=}0.98$, and use learning rate warmup over the first 10\% training steps followed by linear decay to 0.
Our model is implemented in PyTorch~\cite{paszke2019pytorch} and transformers~\cite{wolf-etal-2020-transformers}. 
It is trained for 40 epochs with mixed precision, on 8 NVIDIA V100 GPUs with a batch size of 32 per GPU. The whole training process takes 4 days to complete.

For downstream finetuning,
we use the same training and optimizer configurations except that the default input image size is set to 448 (due to the typically lower resolution of videos compared to images). Since downstream datasets vary in scale and domain, we use task-specific learning rates and training epochs based on validation performance.

\section{Experiments}\label{sec:experiments}
In this section, we evaluate \ModelName~on two popular video-and-language tasks, text-to-video retrieval and video question answering, across six different datasets.
We also provide extensive ablation studies to analyze the key factors that contribute to \ModelName's success.

\subsection{Downstream Tasks}
\noindent\textbf{Text-to-Video Retrieval.} 
$(i)$ \textbf{MSRVTT}~\cite{xu2016msr} contains 10K YouTube videos with 200K descriptions. 
We follow~\cite{yu2018joint,miech2019howto100m}, using 7k train+val videos for training and report results on the 1K test set~\cite{yu2018joint}. 
We also create a local val set by sampling 1K video-caption pairs from unused test videos for our ablation study.
$(ii)$ \textbf{DiDeMo}~\cite{anne2017localizing} contains 10K Flickr videos annotated with 40K sentences.
$(iii)$ \textbf{ActivityNet Captions}~\cite{krishna2017dense} contains 20K YouTube videos annotated with 100K sentences.
The training set contains 10K videos, and we use val1 set with 4.9K videos to report results. 
For MSRVTT, we evaluate standard sentence-to-video retrieval. For DiDeMo and ActivityNet Captions, we follow~\cite{zhang2018cross, liu2019use} to evaluate paragraph-to-video retrieval, where all sentence descriptions for a video are concatenated to form a paragraph for retrieval. We use average recall at K (R@K) and median rank (MdR) to measure performance.

\noindent\textbf{Video Question Answering.} 
$(i)$ \textbf{TGIF-QA}~\cite{jang2017tgif} contains 165K QA pairs on 72K GIF videos. 
We experiment with 3 TGIF-QA tasks: \textit{Repeating Action} and \textit{State Transition} for multiple-choice QA, and \textit{Frame QA} for open-ended QA. 
We leave the \textit{Count} task as future work as it requires directly modeling full-length videos.
$(ii)$ \textbf{MSRVTT-QA}~\cite{xu2017video} is created based on videos and captions in MSRVTT, containing 10K videos and 243K open-ended questions. 
$(iii)$ \textbf{MSRVTT multiple-choice test}~\cite{yu2018joint} is a multiple-choice task with videos as questions, and captions as answers.
Each video contains 5 captions, with only one positive match.
For all the QA tasks, we use standard train/val/test splits and report accuracy to measure performance.

\begin{figure}[!t]
  \centering
  \vspace{-5pt}
  \includegraphics[width=\linewidth]{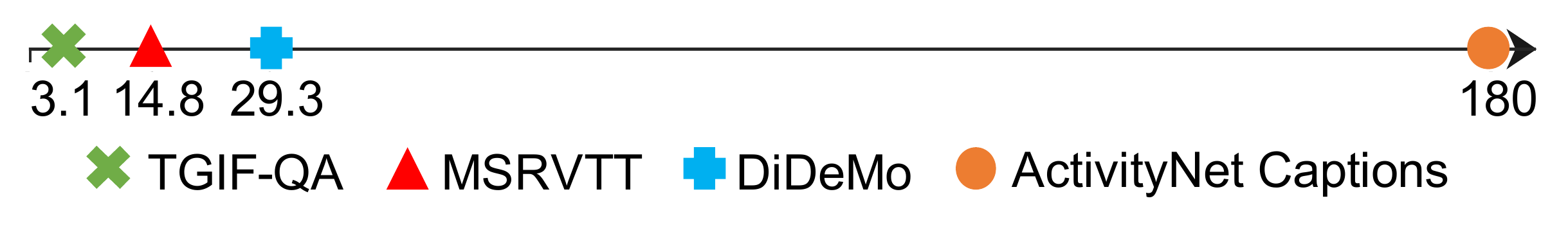}
  \vspace{-15pt}
  \caption{
Average video length in different datasets. 
}
  \label{fig:video_duration}
\end{figure}

Figure~\ref{fig:video_duration} shows a comparison of average video length of evaluated datasets. 
Videos across datasets demonstrate considerable difference in domains and lengths, ranging from 3-second generic-domain GIF videos in TGIF-QA to 180-second human activity videos in ActivityNet Captions.

\begin{table}[!t]
\tablestyle{5pt}{1.1} 
\def\w{20pt} 
\scalebox{1.0}{
\begin{tabular}{c|R{14pt}R{\w}R{\w}R{\w}|c}
\multirow{2}{*}{ $L$ } & \multicolumn{4}{c|}{ MSRVTT Retrieval } & MSRVTT- \\  \cline{2-5}
& R1 & R5 & R10 & MdR & QA Acc. \\
\shline
224 & 6.8 & 24.4 & 35.8 & 20.0 & \bf 35.78 \\
448 & \bf 10.2 & \bf 28.6 & \bf 40.5 & \bf 17.0 & \bf  35.73 \\
768 & \bf 11.0 & 27.8 & \bf 40.9 & \bf 16.0 & \bf 35.73 \\
1000 & 10.0 & \bf 28.4 & 39.4 & 18.0 & 35.19 \\
\end{tabular}
}
\caption{
Impact of \textbf{input image size $L$}. 
}
\label{tab:image_size}
\vspace{-3mm}
\end{table}

\subsection{Analysis of Sparse Sampling}\label{subsec:analysis_sparse_sampling}
We conduct comprehensive ablation studies concerning various aspects of \ModelName's design in this section and Section~\ref{subsec:analysis_pretraining_and_e2e_training}.
If not otherwise stated, we randomly sample a single frame ($N_{train}{=}1$ and $T{=}1$) from full-length videos for training, and use the middle frame ($N_{test}{=}1$) for inference, with input image size $L{=}448$. 
All ablation results are on MSRVTT retrieval local val and MSRVTT-QA val sets.

\paragraph{Do we need larger input image size?} 
We compare models with different input image sizes $L \in \{224, 448, 768, 1000\}$, results shown in Table~\ref{tab:image_size}.
Compared to the model with $L{=}224$, larger input resolution improves performance on the retrieval task, while maintaining a similar performance on the QA task.
The best performance is achieved at around $L{=}448$. Further increasing the resolution does not provide significant performance boost.
\cite{jiang2020defense} shows that increasing input image size from 448 to 1333 always improves image VQA~\cite{antol2015vqa} performance with no sign of saturation, while we observe the performance converges at around 448 for MSRVTT retrieval and QA.
This is potentially because VQA images are typically of higher raw resolution than MSRVTT videos (we are only able to obtain videos at a maximum height of 240 pixels).
We expect higher resolution videos could further improve model performance.

\paragraph{Do we need densely sampled frames?} 
A common practice for video understanding and video+language understanding is to model densely sampled frames from the original video 
(\eg,~\cite{carreira2017quo,Xie_2018_ECCV} sample frames at 25 frames per second).
To understand the impact of using densely sampled frames, we conduct a set of controlled experiments. 
Specifically, we randomly sample a fixed-length 1-second clip from the video, then evenly sample $T {=} \{1, 2, 4, 8, 16\}$ frames within this clip for training. 
For inference, we use the middle clip of the video.
When multiple frames are used (\ie, $T$\textgreater1), we use mean pooling for temporal fusion. 

We also experiment with variants using additional 3D convolutions before mean pooling: ($i$) Conv3D: a standard 3D convolution layer with kernel size 3, stride 1; ($ii$) Conv(2+1)D: a spatial and temporal separable 3D convolution~\cite{tran2018closer,Xie_2018_ECCV}.
Adding 3D convolutions to 2D convolutions essentially leads to a design similar to Top-Heavy S3D architecture~\cite{Xie_2018_ECCV}, which shows better performance than full 3D convolutions on video action recognition and runs faster.

Results are shown in Table~\ref{tab:num_frames_in_training}.
Overall, models that use 3D convolutions perform worse than models that adopt a simple mean pooling. 
For mean pooling, we observe that using two frames provides a notable improvement over using a single frame.
However, models that use more than two frames perform similarly compared to the one using two frames, suggesting that two frames already represent enough local temporal information for the tasks.

\begin{table}[!t]
\tablestyle{5pt}{1.1}
\def\w{20pt} 
\scalebox{1.0}{
\begin{tabular}{c|c|R{12pt}R{\w}R{\w}R{\w}|c}
\multirow{2}{*}{$\mathcal{M}$} & \multirow{2}{*}{$T$} & \multicolumn{4}{c|}{ MSRVTT Retrieval } & MSRVTT- \\ \cline{3-6}
 &  & R1 & R5 & R10 & MdR & QA Acc. \\
\shline
- & 1 & 10.2 & 28.6 & 40.5 & 17.0 & 35.73 \\ 
\hline
\multirow{4}{*}{ Mean Pooling } & 2 & \bf 11.3 & 31.7 & 44.9 & 14.0 & \bf 36.02 \\
& 4 & 10.8 & 30.0 & 43.6 & 14.0  & 35.83 \\
& 8 & 10.6 & \bf 32.5 & \bf 45.0 & \bf 13.0 & 35.69 \\
& 16 & \bf 11.6 & \bf 33.9 & \bf 45.8 & \bf 13.0 & \bf 36.05 \\
\hline
\multirow{2}{*}{ Conv3D } & 2 & 8.7 & 27.3 & 40.2 & 17.0 & 34.85 \\
& 16 & 10.1 & 28.9 & 41.7 & 16.0 & 35.03 \\
\hline
\multirow{2}{*}{ Conv(2+1)D } & 2 & 7.3 & 24.1 & 35.6 & 22.0 & 34.13 \\
& 16 & 9.9 & 27.3 & 39.6 & 17.0 & 33.92 \\
\end{tabular}
}
\caption{
Impact of \textbf{\#frames ($T$) and temporal fusion function ($\mathcal{M}$).}
We use a 1-second clip for all experiments.
}
\label{tab:num_frames_in_training}

\end{table}

\begin{figure}[!t]
  \centering
  \includegraphics[width=\linewidth]{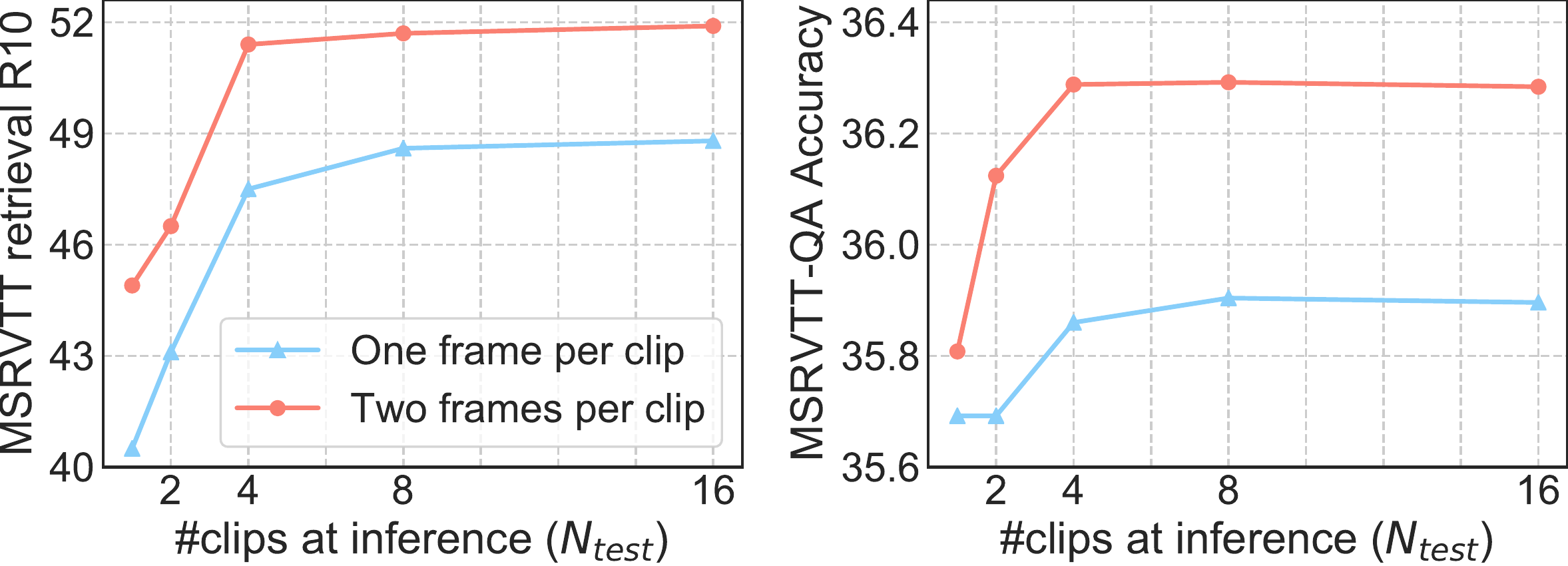}
  \caption{
Impact of \textbf{\#inference clips ($N_{test}$).}
}
  \label{fig:num_clips_at_inferece}
  \vspace{-3mm}
\end{figure}

\paragraph{Do more clips at inference help?} 
At inference, we aggregate prediction scores from multiple densely sampled clips as the final score. 
To show how this strategy affects performance, we evenly sample $N_{test} \in \{1, 2, 4, 8, 16\}$ clips from a video and average their individual predictions at inference. 
For this experiment, we provide two models trained with different numbers of training frames per clip: one with a single frame and the other with two frames.
Both models use a single clip for training.
Results are shown in Figure~\ref{fig:num_clips_at_inferece}.
Adding more clips generally improves performance, especially the first few additions, but after a certain point performance saturates.
For example, in Figure~\ref{fig:num_clips_at_inferece} (\textit{left}), MSRVTT retrieval performance increases substantially as we use two and four clips, compared to using a single clip; then the performance gain gradually becomes marginal.

\begin{table}[!t]
\tablestyle{4pt}{1.1}
\def\w{19pt} 
\scalebox{1.0}{
\begin{tabular}{c|c|R{12pt}R{\w}R{\w}R{\w}|c}
\multirow{2}{*}{$\mathcal{G}$} & \multirow{2}{*}{$N_{train}$} & \multicolumn{4}{c|}{ MSRVTT Retrieval } & MSRVTT- \\ \cline{3-6}
 &  & R1 & R5 & R10 & MdR & QA Acc. \\
\shline
-  & 1 & 12.7 & 34.5 & 48.8 & 11.0 & 36.24 \\
\hline
\multirow{4}{*}{ Mean Pooling } & 2 & 13.3 & 37.1 & 50.6 & 10.0 & 35.94 \\
& 4 & 14.0 & 38.6 & 51.6 & 10.0 &  35.40 \\
& 8 & 13.4 & 36.4 & 49.7 & 11.0 &  35.76 \\
& 16 & 15.2 & \bf 39.4 & \bf 53.1 & \bf 9.0 & 35.33 \\
\hline
\multirow{2}{*}{ Max Pooling } & 2 & 8.5 & 28.7 & 42.2 & 14.0 & \bf 36.41 \\
& 16 & 12.5 & 33.1 & 46.8 & 12.0 & 36.25 \\
\hline
\multirow{2}{*}{ LogSumExp } & 2 & \bf 15.5 & 38.4 & 52.6 & \bf 9.0 & \bf 36.59 \\
& 16 & \bf 17.4 & \bf 41.5 & \bf 55.5 & \bf 8.0 & 36.16 \\
\end{tabular}
}
\caption{
Impact of \textbf{\#training clips ($N_{train}$) and score aggregation function ($\mathcal{G}$).}
All models use $N_{test}{=}$16 clips for inference.
}
\label{tab:num_clips_for_training}
\vspace{-3mm}
\end{table}

\begin{table}[!t]
\tablestyle{2pt}{1.1}
\def\w{19pt}
\scalebox{1.0}{
\begin{tabular}{c|c|R{12pt}R{\w}R{\w}R{\w}|c}
\multirow{2}{*}{Sampling Method} & \multirow{2}{*}{$N_{train}$} & \multicolumn{4}{c|}{ MSRVTT Retrieval } & MSRVTT- \\ \cline{3-6}
 &  & R1 & R5 & R10 & MdR & QA Acc. \\
\shline
\multirow{1}{*}{ Dense Uniform } & 16 & 15.5 & 39.6 & 55.0 & 9.0 & 35.88 \\
\hline
\multirow{3}{*}{ Sparse Random } & 1 & 12.7 & 34.5 & 48.8 & 11.0 & 36.24 \\
 & 2 & 15.5 & 38.4 & 52.6 & 9.0 & 36.59 \\
& 4 & \bf 15.7 & \bf 41.9 & \bf 55.3 & \bf 8.0 & \bf 36.67 \\
\end{tabular}
}
\caption{
\textbf{Sparse random sampling \textit{vs.} dense uniform sampling.}
All models use $N_{test}{=}$16 clips for inference. 
}
\label{tab:dense_uniform_vs_sparse_random}
\vspace{-3mm}
\end{table}

\paragraph{Do more clips in training help?} 
We randomly sample $N_{train}$ clips and aggregate scores from the clips with aggregation function $\mathcal{G}$ as the final score to calculate the training loss.
When multiple clips are used, information from these samples is aggregated through multiple instance learning to maximize the utility of these clips. 
To understand how this strategy affects model performance, we evaluate model variants that use $N_{train} \in \{1, 2, 4, 8, 16\}$ at training.
We also consider 3 different commonly used score aggregation functions for $\mathcal{G}$: mean pooling, max pooling, and LogSumExp~\cite{miech2020end}. In mean pooling and max pooling, the cross-clip pooling is performed over logits, followed by a softmax operator. In LogSumExp, logits from each clip are first fed through an element-wise exponential operator, followed by a cross-clip mean pooling. The aggregated output is further normalized by its own sum to make it eligible as a probability distribution. For simplicity, we always use the same aggregation function for training and inference. 
For a fair comparison, all models use a single frame per clip for training and 16 clips for inference, \ie, $T{=}1$ and $N_{test}{=}16$.

Results are shown in Table~\ref{tab:num_clips_for_training}.
In general, adding more clips helps, and the second added clip gives the most performance gain. 
For example, for models with LogSumExp, $N_{train}{=}2$ improves retrieval R1 score of $N_{train}{=}1$ by 2.8\%, while $N_{train}{=}16$ improves only 1.9\% upon $N_{train}{=}2$, even though it adds much more clips.
As for score aggregation function $\mathcal{G}$, LogSumExp works the best.

\begin{figure}[!t]
  \centering
  \includegraphics[width=\linewidth]{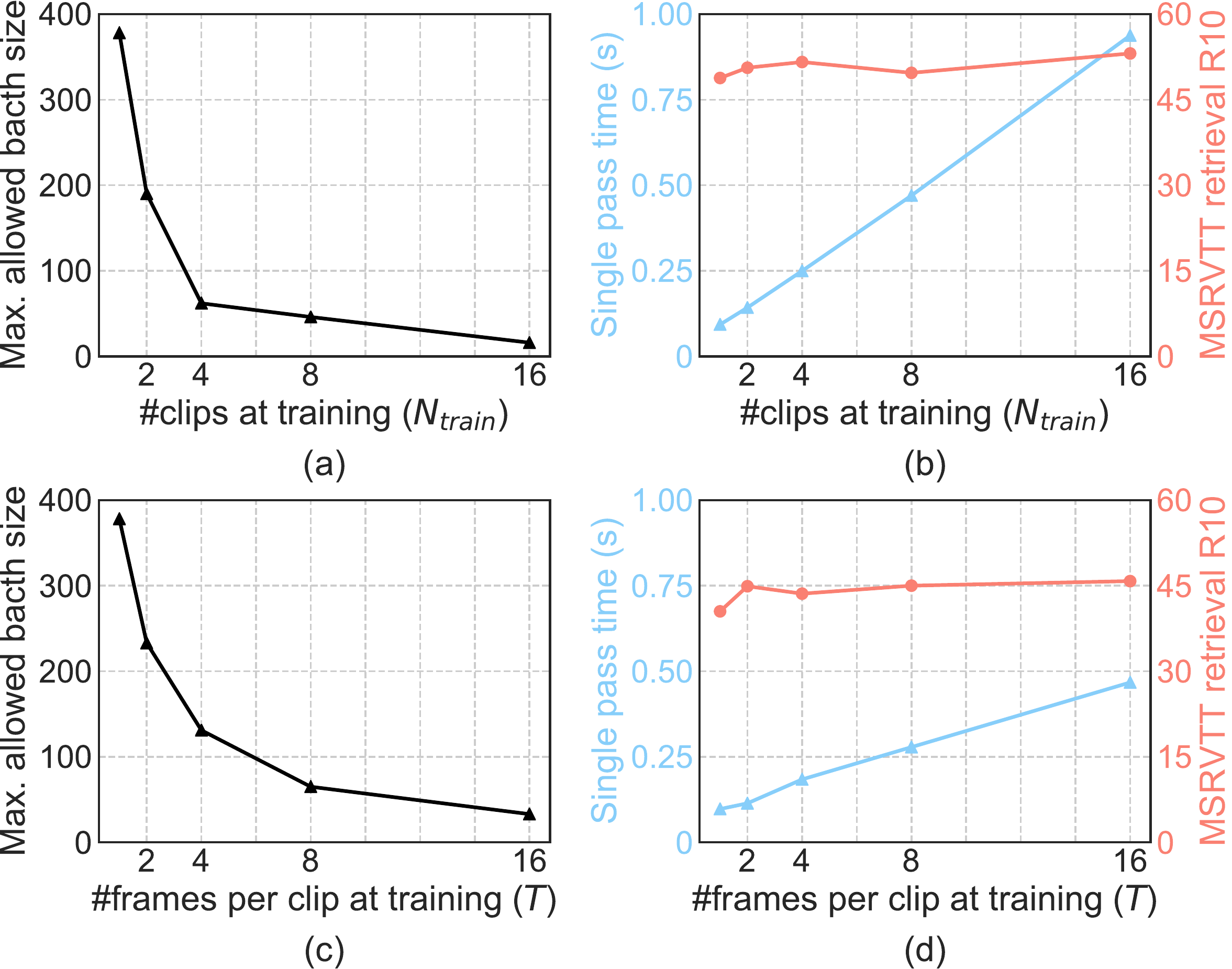}
  \vspace{-3mm}
  \caption{
\textbf{Memory and computation cost comparison} w.r.t. different numbers of clips ($N_{train}$) or frames ($T$) at training. 
\textbf{(a)}: Maximum allowed batch size with fixed $T{=}1$. 
\textbf{(b)}: Time cost for a single forward and backward pass with fixed $T{=}1$, batch size 8. 
\textbf{(c)}: Maximum allowed batch size with fixed $N_{train}{=}1$. 
\textbf{(d)}: Time cost for a single forward and backward pass with fixed $N_{train}{=}1$, batch size 8.
All experiments are conducted on a single NVIDIA V100 GPU with 32GB memory. 
MSRVTT retrieval performance is also added in (b, d) for reference. Best viewed in color.
}
  \label{fig:memory_computation_cost}
  \vspace{-3mm}
\end{figure}

\paragraph{Sparse Random Sampling \textit{vs.} Dense Uniform Sampling.} 
At each training step, \ModelName~randomly samples only a single or a few short clips from a full-length video.
Intuitively, this sparse random sampling strategy can be interpreted as a type of data augmentation where different subsets of clips for a video are used to calculate the loss at different training steps.
To show the effectiveness of this approach, we compare \ModelName~with a variant that uses uniformly sampled dense clips.
Specifically, we use the same \ModelName~architecture as before but always uses 16 uniformly sampled clips for training.
Table~\ref{tab:dense_uniform_vs_sparse_random} shows the comparison. 
Sparse random sampling with only 4 clips outperforms dense uniformly sampling with 16 clips across all metrics in both retrieval and QA tasks.
Meanwhile, using 4 clips is much more memory and computation efficient than using 16 clips, as we show in the next paragraph.
In addition to these two sampling approaches, it is also possible to choose clips using content-based methods such as~\cite{wu2019adaframe}. 
However, this requires an extra non-trivial selection step, and may also remove some of the data augmentation effect brought by random sampling.

\paragraph{Memory and Computation Cost.} 
Figure~\ref{fig:memory_computation_cost} shows a comparison of memory and computation cost \emph{w.r.t.} different numbers of clips ($N_{train}$) or frames ($T$) at training. 
We observe that using more clips or more frames at training considerably increases memory demand and computational cost.
For example, in Figure~\ref{fig:memory_computation_cost}~(\textit{a}), we see that the maximum allowed batch size for a single NVIDIA V100 GPU is 190 when $N_{train}{=}2$, compared to that of 16 when $N_{train}{=}16$.
Meanwhile, in Figure~\ref{fig:memory_computation_cost}~(\textit{b}), we see that the time cost increases almost linearly with $N_{train}$, while the performance improvement on MSRVTT retrieval is less significant.
These comparisons demonstrate the efficiency and effectiveness of the proposed sparse training strategy.

\subsection{Analysis of Pre-training/End-to-end Training}\label{subsec:analysis_pretraining_and_e2e_training}

\begin{table}[!t]
\tablestyle{5pt}{1.1}
\scalebox{1.0}{
\begin{tabular}{l|l|R{12pt}R{11pt}R{11pt}R{20pt}|c}
\multicolumn{2}{c|}{ Weight Initialization } & \multicolumn{4}{c|}{ MSRVTT Retrieval } & MSRVTT- \\  \cline{1-6}
CNN & transformer & R1 & R5 & R10 & MdR & QA Acc. \\
\shline
random & random & 0.3 & 0.4 & 0.9 & 506.0 & 28.05 \\
random & $\text{BERT}_{\textsc{base}}$ & 0.0 & 0.2 & 0.7 & 505.0 & 31.72 \\
TSN, K700 & $\text{BERT}_{\textsc{base}}$ & 5.7 & 22.1 & 33.1 & 23.0 & 35.40 \\
ImageNet & $\text{BERT}_{\textsc{base}}$  & 7.2 & 23.3 & 35.6 & 21.0 & 35.01 \\
grid-feat & $\text{BERT}_{\textsc{base}}$  & 7.4 & 21.0 & 30.7 & 26.0 & 35.27 \\
\hline
\multicolumn{2}{c|}{ image-text pre-training } & \bf 10.2 & \bf 28.6 & \bf 40.5 & \bf 17.0 & \bf 35.73 \\
\end{tabular}
}
\caption{
Impact of \textbf{weight initialization strategy}.
}
\label{tab:weight_init}
\vspace{-3mm}
\end{table}

\paragraph{Impact of Image-text Pre-training.} 
Our model is initialized with image-text pre-training on COCO and Visual Genome Captions, to obtain better-aligned visual and textual representations.
To validate the effectiveness of using image-text pre-training for weight initialization, we also evaluate variants that use other weight initialization strategies. 
Specifically, for CNN, we use weights from random initialization, image classification model pre-trained on ImageNet~\cite{deng2009imagenet}, frame-wise action recognition model TSN~\cite{wang2016temporal,2020mmaction2} pre-trained on Kinetics-700~\cite{smaira2020short,carreira2017quo}, or detection model grid-feat~\cite{jiang2020defense} pre-trained on Visual Genome~\cite{krishna2017visual}. 
For transformer and word embedding layers, we use weights from random initialization or pre-trained $\text{BERT}_{\textsc{base}}$ model~\cite{devlin2018bert}.
For random initialization, we use the default setup in PyTorch~\cite{paszke2019pytorch} and Transformer~\cite{Wolf2019HuggingFacesTS} libraries for CNN and transformer layers, respectively.
Results are summarized in Table~\ref{tab:weight_init}.
We notice that randomly initializing CNN leads to massive performance degradation or even training failure, we hypothesize that it is mostly because of the difficulty of training large models on relatively small datasets (\eg, MSRVTT retrieval train split: 7K videos).
The best performance is achieved using image-text pre-trained weights, clearly indicating the benefit of utilizing image-text pre-training for video-text tasks.

\paragraph{Impact of End-to-End Training.} 
The standard paradigm for video-and-language understanding is to train models with offline extracted features, in which the task objective does not affect the video and text encoding process. 
In this work, we train our model in an end-to-end manner, allowing the model to finetune feature representations by leveraging task supervision. 
In Table~\ref{tab:e2e}, we compare our model with variants that freeze portions of the parameters.
Overall, the best performance is achieved by our model, showing the importance of end-to-end training.
Note that all the models in Table~\ref{tab:e2e} are finetuned from our end-to-end image-text  pre-trained model, which partly resolves the multimodal feature disconnection issue in Section~\ref{sec:intro}. 
Thus, we expect smaller improvement from further end-to-end finetuning.

\paragraph{Main Conclusions} from the analyses in Section~\ref{subsec:analysis_sparse_sampling} and Section~\ref{subsec:analysis_pretraining_and_e2e_training} are summarized as follows:
($i$) Larger input image size helps improve model performance, but the gain diminishes when image size is larger than 448;
($ii$) Sparsely sampling 2 frames from each clip performs on par with dense sampling 16 frames, showing that just one or two frames are sufficient for effective video-and-language training; 
mean-pooling is more effective than 3D Conv when fusing information from different frames;
($iii$) More clips at inference helps improve model performance; aggregation strategy of predictions across clips affects final performance;
($iv$) Sparse (random) sampling is more effective than dense uniform sampling while being more memory and computation efficient;
($v$) Image-text pre-training benefits video-text tasks; 
and ($vi$) End-to-end training improves model performance.

\begin{table}[!t]
\tablestyle{2pt}{1.1}
\def\w{20pt}  
\scalebox{1.0}{
\begin{tabular}{c|c|R{12pt}R{16pt}R{18pt}R{\w}|c}
\multicolumn{2}{c|}{ Parameters Trainable? } & \multicolumn{4}{c|}{ MSRVTT Retrieval } & MSRVTT- \\ \cline{1-6}
\quad\;$\mathcal{F}_{v}$\quad\; & $\mathcal{F}_{l}$ & R1 & R5 & R10 & MdR & QA Acc. \\
\shline
\xmark & \xmark & 8.0 & 27.2 & 38.9 & 17.0 & \bf 35.78 \\
\xmark & \cmark & 9.0 & 27.5 & 39.4 & 18.0 & 35.50 \\
\hline
\cmark & \cmark & \bf 10.2 & \bf 28.6 & \bf 40.5 & \bf 17.0 & 35.73 \\
\end{tabular}
}
\caption{
Impact of \textbf{end-to-end training}.
}
\label{tab:e2e}
\vspace{-3mm}
\end{table}

\begin{table*}[t!]\centering
\subfloat[MSRVTT 1K test set.
\label{tab:main_msrvtt_retrieval}]{\tablestyle{2pt}{1.1}
\def\w{15pt} 
\begin{tabular}{l|R{14pt}R{\w}R{\w}r}
\multicolumn{1}{c|}{Method} & R1 & R5 & R10 & MdR \\
\shline
\demph{HERO~\cite{li2020hero} ASR, PT} & \demph{20.5} & \demph{47.6} & \demph{60.9} & \demph{-} \\
\hline
JSFusion~\cite{yu2018joint} & 10.2 & 31.2 & 43.2 & 13.0 \\
HT~\cite{miech2019howto100m} PT & 14.9 & 40.2 & 52.8 & 9.0 \\
ActBERT~\cite{zhu2020actbert} PT & 16.3 & 42.8 & 56.9 & 10.0 \\
HERO~\cite{li2020hero} PT & 16.8 & 43.4 & \bf 57.7 & - \\
\hline
\ModelName~4\x1 & \bf 19.8 & \bf 45.1 & 57.5 & \bf 7.0 \\
\ModelName~8\x2 & \bf 22.0 & \bf 46.8 & \bf 59.9 & \bf 6.0 \\
\multicolumn{5}{c}{} \\
\end{tabular}
}
\hspace{1mm}
\subfloat[DiDeMo test set.  \label{tab:main_didemo_retrieval}]{\tablestyle{2pt}{1.1}
\def\w{15pt}  
\begin{tabular}{l|R{14pt}R{\w}R{\w}r}
\multicolumn{1}{c|}{Method} & R1 & R5 & R10 & MdR  \\
\shline
\demph{CE~\cite{liu2019use}} & \demph{16.1}	& \demph{41.1} & \demph{-} & \demph{8.3}  \\
\hline
S2VT~\cite{venugopalan2014translating} & 11.9 & 33.6 & - & 13.0 \\
FSE~\cite{zhang2018cross}	& 13.9	& 36.0 & - & 11.0 \\
\hline
\ModelName~4\x1 & \bf 19.9 & \bf 44.5 & \bf 56.7 & \bf 7.0  \\
\ModelName~8\x2 &\bf 20.4	& \bf 48.0	& \bf 60.8	& \bf 6.0\\
\multicolumn{5}{c}{} \\
\multicolumn{5}{c}{} \\
\multicolumn{5}{c}{} \\
\end{tabular}
}
\hspace{1mm}
\subfloat[ActivityNet Captions val1 set. 
\label{tab:main_anet_cap_retrieval}]{\tablestyle{2pt}{1.1}
\def\w{15pt} 
\begin{tabular}{l|R{14pt}R{\w}R{\w}r}
\multicolumn{1}{c|}{Method} & R1 & R5 & R10 & MdR  \\
\shline
\demph{CE~\cite{liu2019use}} & \demph{18.2}	& \demph{47.7} & \demph{-} & \demph{6.0}  \\
\demph{MMT~\cite{gabeur2020multi}} & \demph{22.7} & \demph{54.2} & \demph{93.2} & \demph{5.0}  \\
\demph{MMT~\cite{gabeur2020multi} PT} & \demph{28.7} & \demph{61.4} & \demph{94.5} & \demph{3.3}  \\
\hline
Dense~\cite{krishna2017dense} & 14.0 & 32.0 & - & 34.0 \\
FSE~\cite{zhang2018cross}	& 18.2	& 44.8 & - & 7.0 \\
HSE~\cite{zhang2018cross}	& 20.5	& \bf 49.3 & - & - \\
\hline
\ModelName~4\x2$^*$ & \bf 20.9 & 48.6 & \bf 62.8 & \bf 6.0 \\
\ModelName~4\x2$^*$ ($N_{test}{=}$20)  & \bf 21.3 & \bf 49.0 & \bf 63.5 & \bf 6.0 \\
\end{tabular}
}
\caption{
\textbf{Comparison with state-of-the-art methods on text-to-video retrieval}.
\ModelName~models with different training input sampling methods are denoted by $N_{train}$\x$T$. 
We use $N_{test}{=}16$ if not otherwise stated. 
We gray out models that used features other than appearance and motion for a fair comparison, \eg, CE used appearance, scene, motion, face, audio, OCR, ASR features from 11 different models.
\textit{PT} indicates the model is pre-trained on HowTo100M. * denotes models use 2-second clips instead of the default 1-second clips.
\label{tab:main_retrieval}
}
\vspace{-3mm}
\end{table*}

\begin{table*}[!t]
\subfloat[TGIF-QA test set.
\label{tab:main_tgif_qa}]{\tablestyle{3pt}{1.1}
\def\w{30pt} 
\begin{tabular}{l|C{24pt}C{\w}C{\w}}
\multicolumn{1}{c|}{Method} & Action & Transition & FrameQA \\
\shline
ST-VQA~\cite{jang2017tgif} & 60.8 & 67.1 & 49.3 \\
Co-Memory~\cite{gao2018motion} & 68.2 & 74.3 & 51.5 \\
PSAC~\cite{li2019beyond} & 70.4 & 76.9 & 55.7 \\
Heterogeneous Memory~\cite{fan2019heterogeneous} & 73.9 & 77.8 & 53.8 \\
HCRN~\cite{le2020hierarchical} & 75.0 & 81.4 & 55.9 \\
QueST~\cite{jiang2020divide} & 75.9 & 81.0 & \bf 59.7 \\
\hline
\ModelName~1\x1 ($N_{test}{=}$1) & \bf 82.9 & \bf 87.5 & 59.4 \\
\ModelName~1\x1 & \bf 82.8 & \bf 87.8 & \bf 60.3 \\
\end{tabular}
}
\hspace{2mm}
\subfloat[MRSVTT-QA test set.
\label{tab:main_msrvtt_qa}]{\tablestyle{3pt}{1.1}
\def\w{40pt} 
\begin{tabular}{l|C{30pt}}
\multicolumn{1}{c|}{Method} & Accuracy \\
\shline
ST-VQA~\cite{jang2017tgif} (by \cite{fan2019heterogeneous}) & 30.9  \\
Co-Memory~\cite{gao2018motion} (by \cite{fan2019heterogeneous}) & 32.0 \\
AMU~\cite{xu2017video} & 32.5 \\
Heterogeneous Memory~\cite{fan2019heterogeneous} & 33.0 \\
HCRN~\cite{le2020hierarchical} & 35.6 \\
\hline
\ModelName~4\x1 & \bf 37.0 \\
\ModelName~8\x2 & \bf 37.4 \\
\multicolumn{2}{c}{} \\
\end{tabular}
}
\hspace{2mm}
\subfloat[MRSVTT multiple-choice test. \label{tab:main_msrvtt_multiple_choice_test}]{\tablestyle{3pt}{1.1}
\def\w{40pt} 
\begin{tabular}{l|C{30pt}}
\multicolumn{1}{c|}{Method} & Accuracy \\
\shline
SNUVL~\cite{yu2016video} (by \cite{yu2018joint})  & 65.4 \\
ST-VQA~\cite{jang2017tgif} (by \cite{yu2018joint}) & 66.1 \\
CT-SAN~\cite{yu2017end} (by \cite{yu2018joint}) & 66.4 \\
MLB~\cite{kim2016hadamard} (by \cite{yu2018joint}) & 76.1 \\
JSFusion~\cite{yu2018joint} & 83.4 \\
ActBERT~\cite{zhu2020actbert} PT & 85.7 \\
\hline
\ModelName~4\x1 & \bf 87.9 \\
\ModelName~8\x2 & \bf 88.2 \\
\end{tabular}
}\caption{
\textbf{Comparison with state-of-the-art methods on video question answering}.
}
\label{tab:main_video_qa}
\end{table*}

\subsection{Comparison to State-of-the-art}
For evaluation, we compare \ModelName~with state-of-the-art methods on text-to-video retrieval and video question answering tasks. 
We denote models using different sampling methods at training as \textit{\ModelName~$N_{train}$\x$T$}, (randomly sample $N_{train}$ 1-second clips for training, each contains $T$ uniformly sampled frames of size $L{=}448$).
We use LogSumExp to aggregate scores from multiple clips.
At inference time, if not otherwise stated, we aggregate scores from $N_{test}{=}$16 uniformly sampled clips.

\paragraph{Text-to-Video Retrieval.}
Table~\ref{tab:main_retrieval} summarizes results on text-to-video retrieval. 
In Table~\ref{tab:main_msrvtt_retrieval}, \ModelName~achieves significant performance gain over existing methods on MSRVTT retrieval, including HT~\cite{miech2019howto100m}, ActBERT~\cite{zhu2020actbert}, and HERO~\cite{li2020hero}, which are pre-trained on 136M clip-caption pairs from HowTo100M~\cite{miech2019howto100m}. 
Under a fair comparison, \ModelName~4\x1 outperforms HERO~\cite{li2020hero} by 3.0\% on R@1.
Note that HERO uses SlowFast~\cite{feichtenhofer2019slowfast} features extracted from full-length videos at a very dense frame rate of 21 FPS (\ie, on average 310 frames per video for MSRVTT), while \ModelName~4\x1 uses only 4 randomly sampled frames.
When more frames are used, \ModelName~8\x2 achieves even higher performance, surpassing HERO by 5.2\%.
Compared to the HERO ASR model that uses extra input from Automatic Speech Recognition (ASR), \ModelName~still obtains 1.5\% higher R1 score.

Similarly, on DiDeMo and ActivityNet Captions paragraph-to-video retrieval tasks (Table~\ref{tab:main_didemo_retrieval} and Table~\ref{tab:main_anet_cap_retrieval}), we notice our best \ModelName~models outputform CE~\cite{liu2019use} by 4.3\% and 3.1\% on R1, respectively, despite CE's use of appearance, scene, motion, face, audio, OCR, ASR features densely extracted from 11 different models.
ActivityNet Caption videos are on average 180-second long. 
In Table~\ref{tab:main_anet_cap_retrieval} we show \ModelName~performs competitively with existing methods that model long-range relations in this dataset. 
Especially, \ModelName~obtains 0.8\% higher R1 than HSE~\cite{zhang2018cross} and is competitive compared to MMT~\cite{gabeur2020multi} that uses extra audio features\footnote{\cite{gabeur2020multi} shows that adding audio features greatly improves performance.}, even though \ModelName~4\x2$^*$ ($N_{test}{=}20$) samples only 8-second clips from 180-second videos at each training step, and uses only 40-second content for inference.
We expect \ModelName's performance to be further improved by sampling more clips during training and inference.
Meanwhile, we also encourage future work to explore combining extra input signals, such as audio, into the \ModelName~framework for better performance.

\paragraph{Video Question Answering.}
Table~\ref{tab:main_video_qa} shows evaluation results on video question answering. 
Across all three tasks, \ModelName~achieves significant performance gain.
In Table~\ref{tab:main_tgif_qa}, \ModelName~1\x1 outperforms prior state-of-the-art QueST~\cite{jiang2020divide} by 6.9\%, 6.8\%, and 0.6\% on TGIF-QA Action, Transition, and FrameQA tasks, respectively.
This is especially surprising considering \ModelName~1\x1 uses only a single randomly sampled frame from the videos at each training step, while QueST uses 10 uniformly sampled frames.
Moreover, when using only a single frame (the middle frames of the videos) for inference, \ModelName~1\x1 ($N_{test}{=}1$) already far outperforms QueST on Action and Transition tasks, and is on par with QueST on FrameQA.
In Table~\ref{tab:main_msrvtt_qa}, \ModelName~4\x1 outperforms HCRN~\cite{le2020hierarchical} by 1.4\% on MSRVTT-QA.
Note that HCRN adopts a sophisticated hierarchical relation modeling network over the entire video sequence of 24 clips at training time, while we use only four randomly sampled frames. 
Using more frames further increases this performance gap to 1.8\%.
Table~\ref{tab:main_msrvtt_multiple_choice_test} shows \ModelName~8\x2 improves ActBERT~\cite{zhu2020actbert} model pre-trained on HowTo100M by 2.5\%, on MSRVTT multiple choice test task.

\section{Conclusion}\label{sec:conclusion}
We present \ModelName, a generic framework for end-to-end video-and-language learning, which adopts sparse sampling to use only a few sampled short clips from the videos at each training step.
Experiments across diverse tasks show that \ModelName~outperforms (or is on par with) state-of-the-art methods with densely sampled offline features, suggesting that the ``\textit{less is more}" principle is highly effective in practice.
Comprehensive ablation studies reveal several key factors that lead to this success, including sparse sampling, end-to-end training, and image-text pre-training.

\smallskip
\noindent
{\bf Acknowledgements:} This research was partially done when Jie was an intern with Microsoft. He was later supported at UNC by NSF Award \#1562098, DARPA KAIROS Grant \#FA8750-19-2-1004, ARO-YIP Award \#W911NF-18-1-0336, and Microsoft Investigator Fellowship. The views contained in this article are those of the authors and not of the funding agency.

\appendix
\section{Additional Experiments}

\paragraph{Visual Question Answering.} As \ModelName~is designed based on 2D CNN, and is pre-trained on image-text corpus, it is also directly applicable to image-text downstream tasks, such as image-based question answering. 
We show \ModelName's performance on VQA 2.0 dataset~\cite{goyal2017making} in Table~\ref{tab:main_vqa}.
The model is finetuned from the image-text pre-trained weights on 8GPUs for 13 epochs,  with batch size 32 and learning rate 5e-5.
\ModelName~shows a reasonable performance compared to the strong pre-training baselines.
Note that \ModelName~uses grid features~\cite{jiang2020defense,huang2020pixel} instead of the commonly used region features, which is much more computation efficient, \eg, extracting grid features is around 80\x~faster than extracting region features according to the computation time reported in~\cite{jiang2020defense}.

\begin{table}[t]
\tablestyle{3pt}{1.1}
\def\w{40pt}  
\begin{tabular}{l|c|cc}
\multicolumn{1}{c|}{Method} & feature & test-dev & test-std \\
\shline
BUTD~\cite{anderson2018bottom} & R & 65.32 & 65.67\\
grid-feat~\cite{jiang2020defense} & G & 66.47 & -\\
\hline
ViLBERT~\cite{lu2019vilbert} & R & 70.55 & 70.92 \\
VL-BERT~\cite{su2019vl} & R & 71.16 & - \\
Pixel-BERT~\cite{huang2020pixel} & G & 71.35 & 71.42 \\
LXMERT~\cite{tan2019lxmert} & R & 72.42 & 72.54 \\
UNITER~\cite{anderson2018bottom} & R & 72.70 & 72.91\\
Oscar~\cite{li2020oscar} & R & 73.16 & 73.44 \\
\hline
\ModelName~1\x1 & G & 69.08 & 69.43\\
\end{tabular}
\caption{
\textbf{Comparison with state-of-the-art methods on VQA}. 
\textit{G} stands for grid features, \textit{R} stands for region features.
}
\label{tab:main_vqa}
\end{table}

\section{Downstream Task Adaptation}
Our \ModelName~is quite generic, once trained, it can be easily adopted and transferred for various downstream tasks. 
In particular, in this work, we focus on text-to-video retrieval and video question answering.

\paragraph{Text-to-video Retrieval.} 
We use a two-layer MLP with the last layer \texttt{[CLS]} token hidden state for a two way (\ie, matched or not matched) classification for retrieval.
We use LogSumExp loss for training. Denote the two-way classification logit output for clip $\tau_i$ from the video associated with the $j$-$th$ example as $\vg_{\tau_i}^{(j)} \in \mathbb{R}^2$, where $i=1,\ldots, N_{train}$ for training ($i=1,\ldots, N_{test}$ for inference; see Section 3 of the main paper). 
The LogSumExp prediction $\vp^{(j)} \in \mathbb{R}^2$ is defined as:
\begin{equation}\label{eq:supp1}
\vp^{(j)}=\frac{\sum_{i=1}^{N_{train}}{e^{\vg_{\tau_i}^{(j)}}}}{\text{sum}(\sum_{i=1}^{N_{train}}{e^{\vg_{\tau_i}^{(j)}}})}.
\end{equation}

\noindent
We then use a negative log likelihood loss for training:
\begin{equation}\label{eq:supp2}
L = - \frac{1}{|\mathcal{D}|} \sum_{j=1}^{|\mathcal{D}|}\mathrm{log}\vp^{(j)}[y_j],
\end{equation}
where $\mathcal{D}$ is the dataset, $y_j$ is the index of the ground-truth answer for the $j$-$th$ example.

We conduct experiments on three popular text-to-video retrieval datasets, MSRVTT~\cite{xu2016msr}, DiDeMo~\cite{anne2017localizing}, and ActivityNet Captions~\cite{krishna2017dense}.
Table~\ref{tab:training_details_text_to_video} shows the training details for models on each of the datasets.

\begin{table}[h]
\tablestyle{4pt}{1.1} 
\def\w{20pt}  
\scalebox{1.0}{
\begin{tabular}{l|ccc}
Dataset & \#Epochs & Bsz \x Grad-Accu \x \#GPUs  & LR \\
\shline
MSRVTT & 20 & 16\x1\x8 & 5e-5 \\
DiDeMo & 20 & 8\x4\x8 & 5e-5\\
ActivityNet Captions & 80 & 16\x2\x8 & 5e-5\\
\end{tabular}
}
\caption{
\textbf{Training details for text-to-video retrieval tasks}. Bsz is short for batch size. Grad-Accu stands for gradient accumulation steps. LR means initial learning rate.
}
\label{tab:training_details_text_to_video}
\end{table}

\paragraph{Video Question Answering.} 
Similar to text-to-video retrieval task, we take the last layer \texttt{[CLS]} token hidden state through a two-layer MLP for classification.
We use LogSumExp to aggregate prediction from multiple clips to calculate loss. The formulation of LogSumExp loss is simlar to Equation~\ref{eq:supp1} except that the dimension of $\vg_{\tau_i}$ equals to the number of answer candidates.

We conduct experiments on three video QA datasets, TGIF-QA~\cite{jang2017tgif}, MSRVTT-QA~\cite{xu2017video}, and MSRVTT MC Test~\cite{yu2018joint}.
For TGIF-QA, we evaluate three sub-tasks, \ie, Action, Transition, and FrameQA. We train a separate model for each of the evaluated TGIF-QA tasks.
For MSRVTT MC Test, as it uses the same training set as the MSRVTT retrieval task, we directly use the trained retrieval model to rank the five candidate answers.
Table~\ref{tab:training_details_text_to_video} shows the training details for models on TGIF-QA tasks and MSRVTT-QA.

\begin{table}[h]
\tablestyle{4pt}{1.1} 
\def\w{20pt} 
\scalebox{1.0}{
\begin{tabular}{l|ccc}
Dataset & \#Epochs & Bsz\x Grad-Accu \x \#GPUs & LR \\
\shline
TGIF-QA Action & 55 & 32\x1\x8 & 1e-4 \\
TGIF-QA Transition & 15 & 32\x1\x8 & 1e-4 \\
TGIF-QA FrameQA & 15 & 32\x1\x8 & 1e-4 \\
MSRVTT-QA & 10 & 16\x1\x4 & 5e-5 \\
\end{tabular}
}
\caption{
\textbf{Training details for video question answering tasks}. Bsz is short for batch size.
Grad-Accu stands for gradient accumulation steps. 
LR means initial learning rate.
}
\label{tab:training_details_video_qa}
\end{table}

{\small
\bibliographystyle{ieee_fullname}
\bibliography{egbib}

\begin{thebibliography}{10}\itemsep=-1pt

\bibitem{anderson2018bottom}
Peter Anderson, Xiaodong He, Chris Buehler, Damien Teney, Mark Johnson, Stephen
  Gould, and Lei Zhang.
\newblock Bottom-up and top-down attention for image captioning and visual
  question answering.
\newblock In {\em CVPR}, 2018.

\bibitem{anne2017localizing}
Lisa Anne~Hendricks, Oliver Wang, Eli Shechtman, Josef Sivic, Trevor Darrell,
  and Bryan Russell.
\newblock Localizing moments in video with natural language.
\newblock In {\em ICCV}, 2017.

\bibitem{antol2015vqa}
Stanislaw Antol, Aishwarya Agrawal, Jiasen Lu, Margaret Mitchell, Dhruv Batra,
  C Lawrence~Zitnick, and Devi Parikh.
\newblock Vqa: Visual question answering.
\newblock In {\em ICCV}, 2015.

\bibitem{carreira2017quo}
Joao Carreira and Andrew Zisserman.
\newblock Quo vadis, action recognition? a new model and the kinetics dataset.
\newblock In {\em CVPR}, 2017.

\bibitem{chen2015microsoft}
Xinlei Chen, Hao Fang, Tsung-Yi Lin, Ramakrishna Vedantam, Saurabh Gupta, Piotr
  Doll{\'a}r, and C~Lawrence Zitnick.
\newblock Microsoft coco captions: Data collection and evaluation server.
\newblock {\em arXiv}, 2015.

\bibitem{chen2019uniter}
Yen-Chun Chen, Linjie Li, Licheng Yu, Ahmed~El Kholy, Faisal Ahmed, Zhe Gan, Yu
  Cheng, and Jingjing Liu.
\newblock Uniter: Learning universal image-text representations.
\newblock In {\em ECCV}, 2020.

\bibitem{cho2021unifying}
Jaemin Cho, Jie Lei, Hao Tan, and Mohit Bansal.
\newblock Unifying vision-and-language tasks via text generation.
\newblock {\em arXiv}, 2021.

\bibitem{clark2020electra}
Kevin Clark, Minh-Thang Luong, Quoc~V Le, and Christopher~D Manning.
\newblock Electra: Pre-training text encoders as discriminators rather than
  generators.
\newblock In {\em ICLR}, 2020.

\bibitem{2020mmaction2}
MMAction2 Contributors.
\newblock Openmmlab's next generation video understanding toolbox and
  benchmark.
\newblock \url{https://github.com/open-mmlab/mmaction2}, 2020.

\bibitem{deng2009imagenet}
Jia Deng, Wei Dong, Richard Socher, Li-Jia Li, Kai Li, and Li Fei-Fei.
\newblock Imagenet: A large-scale hierarchical image database.
\newblock In {\em CVPR}, 2009.

\bibitem{devlin2018bert}
Jacob Devlin, Ming-Wei Chang, Kenton Lee, and Kristina Toutanova.
\newblock Bert: Pre-training of deep bidirectional transformers for language
  understanding.
\newblock In {\em NAACL}, 2019.

\bibitem{fan2019heterogeneous}
Chenyou Fan, Xiaofan Zhang, Shu Zhang, Wensheng Wang, Chi Zhang, and Heng
  Huang.
\newblock Heterogeneous memory enhanced multimodal attention model for video
  question answering.
\newblock In {\em CVPR}, 2019.

\bibitem{feichtenhofer2020x3d}
Christoph Feichtenhofer.
\newblock X3d: Expanding architectures for efficient video recognition.
\newblock In {\em CVPR}, 2020.

\bibitem{feichtenhofer2019slowfast}
Christoph Feichtenhofer, Haoqi Fan, Jitendra Malik, and Kaiming He.
\newblock Slowfast networks for video recognition.
\newblock In {\em ICCV}, 2019.

\bibitem{gabeur2020multi}
Valentin Gabeur, Chen Sun, Karteek Alahari, and Cordelia Schmid.
\newblock Multi-modal transformer for video retrieval.
\newblock In {\em ECCV}, 2020.

\bibitem{gan2020large}
Zhe Gan, Yen-Chun Chen, Linjie Li, Chen Zhu, Yu Cheng, and Jingjing Liu.
\newblock Large-scale adversarial training for vision-and-language
  representation learning.
\newblock In {\em NeurIPS}, 2020.

\bibitem{gao2018motion}
Jiyang Gao, Runzhou Ge, Kan Chen, and Ram Nevatia.
\newblock Motion-appearance co-memory networks for video question answering.
\newblock In {\em CVPR}, 2018.

\bibitem{gao2017tall}
Jiyang Gao, Chen Sun, Zhenheng Yang, and Ram Nevatia.
\newblock Tall: Temporal activity localization via language query.
\newblock In {\em ICCV}, 2017.

\bibitem{goyal2017making}
Yash Goyal, Tejas Khot, Douglas Summers-Stay, Dhruv Batra, and Devi Parikh.
\newblock Making the v in vqa matter: Elevating the role of image understanding
  in visual question answering.
\newblock In {\em CVPR}, 2017.

\bibitem{gurari2018vizwiz}
Danna Gurari, Qing Li, Abigale~J Stangl, Anhong Guo, Chi Lin, Kristen Grauman,
  Jiebo Luo, and Jeffrey~P Bigham.
\newblock Vizwiz grand challenge: Answering visual questions from blind people.
\newblock In {\em CVPR}, 2018.

\bibitem{he2016deep}
Kaiming He, Xiangyu Zhang, Shaoqing Ren, and Jian Sun.
\newblock Deep residual learning for image recognition.
\newblock In {\em CVPR}, 2016.

\bibitem{huang2020pixel}
Zhicheng Huang, Zhaoyang Zeng, Bei Liu, Dongmei Fu, and Jianlong Fu.
\newblock Pixel-bert: Aligning image pixels with text by deep multi-modal
  transformers.
\newblock {\em arXiv}, 2020.

\bibitem{jang2017tgif}
Yunseok Jang, Yale Song, Youngjae Yu, Youngjin Kim, and Gunhee Kim.
\newblock Tgif-qa: Toward spatio-temporal reasoning in visual question
  answering.
\newblock In {\em CVPR}, 2017.

\bibitem{jiang2020defense}
Huaizu Jiang, Ishan Misra, Marcus Rohrbach, Erik Learned-Miller, and Xinlei
  Chen.
\newblock In defense of grid features for visual question answering.
\newblock In {\em CVPR}, 2020.

\bibitem{jiang2020divide}
Jianwen Jiang, Ziqiang Chen, Haojie Lin, Xibin Zhao, and Yue Gao.
\newblock Divide and conquer: Question-guided spatio-temporal contextual
  attention for video question answering.
\newblock In {\em AAAI}, 2020.

\bibitem{kay2017kinetics}
Will Kay, Joao Carreira, Karen Simonyan, Brian Zhang, Chloe Hillier, Sudheendra
  Vijayanarasimhan, Fabio Viola, Tim Green, Trevor Back, Paul Natsev, et~al.
\newblock The kinetics human action video dataset.
\newblock {\em arXiv}, 2017.

\bibitem{kim2016hadamard}
Jin-Hwa Kim, Kyoung-Woon On, Woosang Lim, Jeonghee Kim, Jung-Woo Ha, and
  Byoung-Tak Zhang.
\newblock Hadamard product for low-rank bilinear pooling.
\newblock In {\em ICLR}, 2016.

\bibitem{krishna2017dense}
Ranjay Krishna, Kenji Hata, Frederic Ren, Li Fei-Fei, and Juan Carlos~Niebles.
\newblock Dense-captioning events in videos.
\newblock In {\em ICCV}, 2017.

\bibitem{krishna2017visual}
Ranjay Krishna, Yuke Zhu, Oliver Groth, Justin Johnson, Kenji Hata, Joshua
  Kravitz, Stephanie Chen, Yannis Kalantidis, Li-Jia Li, David~A Shamma, et~al.
\newblock Visual genome: Connecting language and vision using crowdsourced
  dense image annotations.
\newblock {\em IJCV}, 2017.

\bibitem{lan2019albert}
Zhenzhong Lan, Mingda Chen, Sebastian Goodman, Kevin Gimpel, Piyush Sharma, and
  Radu Soricut.
\newblock Albert: A lite bert for self-supervised learning of language
  representations.
\newblock In {\em ICLR}, 2020.

\bibitem{le2020hierarchical}
Thao~Minh Le, Vuong Le, Svetha Venkatesh, and Truyen Tran.
\newblock Hierarchical conditional relation networks for video question
  answering.
\newblock In {\em CVPR}, 2020.

\bibitem{lei2020mart}
Jie Lei, Liwei Wang, Yelong Shen, Dong Yu, Tamara~L Berg, and Mohit Bansal.
\newblock Mart: Memory-augmented recurrent transformer for coherent video
  paragraph captioning.
\newblock In {\em ACL}, 2020.

\bibitem{lei2018tvqa}
Jie Lei, Licheng Yu, Mohit Bansal, and Tamara~L Berg.
\newblock Tvqa: Localized, compositional video question answering.
\newblock In {\em EMNLP}, 2018.

\bibitem{lei2019tvqa+}
Jie Lei, Licheng Yu, Tamara~L Berg, and Mohit Bansal.
\newblock Tvqa+: Spatio-temporal grounding for video question answering.
\newblock In {\em ACL}, 2020.

\bibitem{lei2020tvr}
Jie Lei, Licheng Yu, Tamara~L Berg, and Mohit Bansal.
\newblock Tvr: A large-scale dataset for video-subtitle moment retrieval.
\newblock In {\em ECCV}, 2020.

\bibitem{li2020unicoder}
Gen Li, Nan Duan, Yuejian Fang, Ming Gong, Daxin Jiang, and Ming Zhou.
\newblock Unicoder-vl: A universal encoder for vision and language by
  cross-modal pre-training.
\newblock In {\em AAAI}, 2020.

\bibitem{li2020hero}
Linjie Li, Yen-Chun Chen, Yu Cheng, Zhe Gan, Licheng Yu, and Jingjing Liu.
\newblock Hero: Hierarchical encoder for video+ language omni-representation
  pre-training.
\newblock In {\em EMNLP}, 2020.

\bibitem{li2019beyond}
Xiangpeng Li, Jingkuan Song, Lianli Gao, Xianglong Liu, Wenbing Huang, Xiangnan
  He, and Chuang Gan.
\newblock Beyond rnns: Positional self-attention with co-attention for video
  question answering.
\newblock In {\em AAAI}, 2019.

\bibitem{li2020oscar}
Xiujun Li, Xi Yin, Chunyuan Li, Pengchuan Zhang, Xiaowei Hu, Lei Zhang, Lijuan
  Wang, Houdong Hu, Li Dong, Furu Wei, et~al.
\newblock Oscar: Object-semantics aligned pre-training for vision-language
  tasks.
\newblock In {\em ECCV}, 2020.

\bibitem{li2016tgif}
Yuncheng Li, Yale Song, Liangliang Cao, Joel Tetreault, Larry Goldberg,
  Alejandro Jaimes, and Jiebo Luo.
\newblock Tgif: A new dataset and benchmark on animated gif description.
\newblock In {\em CVPR}, 2016.

\bibitem{liu2019use}
Yang Liu, Samuel Albanie, Arsha Nagrani, and Andrew Zisserman.
\newblock Use what you have: Video retrieval using representations from
  collaborative experts.
\newblock In {\em BMVC}, 2020.

\bibitem{liu2019roberta}
Yinhan Liu, Myle Ott, Naman Goyal, Jingfei Du, Mandar Joshi, Danqi Chen, Omer
  Levy, Mike Lewis, Luke Zettlemoyer, and Veselin Stoyanov.
\newblock Roberta: A robustly optimized bert pretraining approach.
\newblock {\em arXiv}, 2019.

\bibitem{loshchilov2017decoupled}
Ilya Loshchilov and Frank Hutter.
\newblock Decoupled weight decay regularization.
\newblock In {\em ICLR}, 2019.

\bibitem{lu2019vilbert}
Jiasen Lu, Dhruv Batra, Devi Parikh, and Stefan Lee.
\newblock Vilbert: Pretraining task-agnostic visiolinguistic representations
  for vision-and-language tasks.
\newblock In {\em NeurIPS}, 2019.

\bibitem{miech2020end}
Antoine Miech, Jean-Baptiste Alayrac, Lucas Smaira, Ivan Laptev, Josef Sivic,
  and Andrew Zisserman.
\newblock End-to-end learning of visual representations from uncurated
  instructional videos.
\newblock In {\em CVPR}, 2020.

\bibitem{miech2019howto100m}
Antoine Miech, Dimitri Zhukov, Jean-Baptiste Alayrac, Makarand Tapaswi, Ivan
  Laptev, and Josef Sivic.
\newblock Howto100m: Learning a text-video embedding by watching hundred
  million narrated video clips.
\newblock In {\em ICCV}, 2019.

\bibitem{mikolov2013distributed}
Tomas Mikolov, Ilya Sutskever, Kai Chen, Greg~S Corrado, and Jeff Dean.
\newblock Distributed representations of words and phrases and their
  compositionality.
\newblock In {\em NeurIPS}, 2013.

\bibitem{nguyen2020revisiting}
Duy-Kien Nguyen, Vedanuj Goswami, and Xinlei Chen.
\newblock Revisiting modulated convolutions for visual counting and beyond.
\newblock {\em arXiv}, 2020.

\bibitem{paszke2019pytorch}
Adam Paszke, Sam Gross, Francisco Massa, Adam Lerer, James Bradbury, Gregory
  Chanan, Trevor Killeen, Zeming Lin, Natalia Gimelshein, Luca Antiga, Alban
  Desmaison, Andreas Kopf, Edward Yang, Zachary DeVito, Martin Raison, Alykhan
  Tejani, Sasank Chilamkurthy, Benoit Steiner, Lu Fang, Junjie Bai, and Soumith
  Chintala.
\newblock Pytorch: An imperative style, high-performance deep learning library.
\newblock In {\em NeurIPS}, 2019.

\bibitem{pennington2014glove}
Jeffrey Pennington, Richard Socher, and Christopher~D Manning.
\newblock Glove: Global vectors for word representation.
\newblock In {\em EMNLP}, 2014.

\bibitem{qiu2017learning}
Zhaofan Qiu, Ting Yao, and Tao Mei.
\newblock Learning spatio-temporal representation with pseudo-3d residual
  networks.
\newblock In {\em CVPR}, 2017.

\bibitem{raffel2020exploring}
Colin Raffel, Noam Shazeer, Adam Roberts, Katherine Lee, Sharan Narang, Michael
  Matena, Yanqi Zhou, Wei Li, and Peter~J Liu.
\newblock Exploring the limits of transfer learning with a unified text-to-text
  transformer.
\newblock {\em JMLR}, 2020.

\bibitem{ren2015faster}
Shaoqing Ren, Kaiming He, Ross Girshick, and Jian Sun.
\newblock Faster r-cnn: Towards real-time object detection with region proposal
  networks.
\newblock In {\em NeurIPS}, 2015.

\bibitem{rohrbach2015dataset}
Anna Rohrbach, Marcus Rohrbach, Niket Tandon, and Bernt Schiele.
\newblock A dataset for movie description.
\newblock In {\em CVPR}, 2015.

\bibitem{simonyan2014two}
Karen Simonyan and Andrew Zisserman.
\newblock Two-stream convolutional networks for action recognition in videos.
\newblock In {\em NeurIPS}, 2014.

\bibitem{simonyan2014very}
Karen Simonyan and Andrew Zisserman.
\newblock Very deep convolutional networks for large-scale image recognition.
\newblock In {\em ICLR}, 2015.

\bibitem{smaira2020short}
Lucas Smaira, Jo{\~a}o Carreira, Eric Noland, Ellen Clancy, Amy Wu, and Andrew
  Zisserman.
\newblock A short note on the kinetics-700-2020 human action dataset.
\newblock {\em arXiv}, 2020.

\bibitem{su2019vl}
Weijie Su, Xizhou Zhu, Yue Cao, Bin Li, Lewei Lu, Furu Wei, and Jifeng Dai.
\newblock Vl-bert: Pre-training of generic visual-linguistic representations.
\newblock In {\em ICLR}, 2020.

\bibitem{sun2019videobert}
Chen Sun, Austin Myers, Carl Vondrick, Kevin Murphy, and Cordelia Schmid.
\newblock Videobert: A joint model for video and language representation
  learning.
\newblock In {\em ICCV}, 2019.

\bibitem{szegedy2015going}
Christian Szegedy, Wei Liu, Yangqing Jia, Pierre Sermanet, Scott Reed, Dragomir
  Anguelov, Dumitru Erhan, Vincent Vanhoucke, and Andrew Rabinovich.
\newblock Going deeper with convolutions.
\newblock In {\em CVPR}, 2015.

\bibitem{tan2019lxmert}
Hao Tan and Mohit Bansal.
\newblock Lxmert: Learning cross-modality encoder representations from
  transformers.
\newblock In {\em EMNLP}, 2019.

\bibitem{tran2015learning}
Du Tran, Lubomir Bourdev, Rob Fergus, Lorenzo Torresani, and Manohar Paluri.
\newblock Learning spatiotemporal features with 3d convolutional networks.
\newblock In {\em ICCV}, 2015.

\bibitem{tran2018closer}
Du Tran, Heng Wang, Lorenzo Torresani, Jamie Ray, Yann LeCun, and Manohar
  Paluri.
\newblock A closer look at spatiotemporal convolutions for action recognition.
\newblock In {\em CVPR}, 2018.

\bibitem{vaswani2017attention}
Ashish Vaswani, Noam Shazeer, Niki Parmar, Jakob Uszkoreit, Llion Jones,
  Aidan~N Gomez, {\L}ukasz Kaiser, and Illia Polosukhin.
\newblock Attention is all you need.
\newblock In {\em NeurIPS}, 2017.

\bibitem{venugopalan2014translating}
Subhashini Venugopalan, Huijuan Xu, Jeff Donahue, Marcus Rohrbach, Raymond
  Mooney, and Kate Saenko.
\newblock Translating videos to natural language using deep recurrent neural
  networks.
\newblock In {\em NAACL}, 2015.

\bibitem{wang2016temporal}
Limin Wang, Yuanjun Xiong, Zhe Wang, Yu Qiao, Dahua Lin, Xiaoou Tang, and Luc
  Van~Gool.
\newblock Temporal segment networks: Towards good practices for deep action
  recognition.
\newblock In {\em ECCV}, 2016.

\bibitem{wang2018non}
Xiaolong Wang, Ross Girshick, Abhinav Gupta, and Kaiming He.
\newblock Non-local neural networks.
\newblock In {\em CVPR}, 2018.

\bibitem{Wolf2019HuggingFacesTS}
Thomas Wolf, Lysandre Debut, Victor Sanh, Julien Chaumond, Clement Delangue,
  Anthony Moi, Pierric Cistac, Tim Rault, Rémi Louf, Morgan Funtowicz, Joe
  Davison, Sam Shleifer, Patrick von Platen, Clara Ma, Yacine Jernite, Julien
  Plu, Canwen Xu, Teven~Le Scao, Sylvain Gugger, Mariama Drame, Quentin Lhoest,
  and Alexander~M. Rush.
\newblock Huggingface's transformers: State-of-the-art natural language
  processing.
\newblock {\em arXiv}, 2019.

\bibitem{wolf-etal-2020-transformers}
Thomas Wolf, Lysandre Debut, Victor Sanh, Julien Chaumond, Clement Delangue,
  Anthony Moi, Pierric Cistac, Tim Rault, Rémi Louf, Morgan Funtowicz, Joe
  Davison, Sam Shleifer, Patrick von Platen, Clara Ma, Yacine Jernite, Julien
  Plu, Canwen Xu, Teven~Le Scao, Sylvain Gugger, Mariama Drame, Quentin Lhoest,
  and Alexander~M. Rush.
\newblock Transformers: State-of-the-art natural language processing.
\newblock In {\em EMNLP: System Demonstrations}, 2020.

\bibitem{wu2015deep}
Jiajun Wu, Yinan Yu, Chang Huang, and Kai Yu.
\newblock Deep multiple instance learning for image classification and
  auto-annotation.
\newblock In {\em CVPR}, 2015.

\bibitem{wu2016google}
Yonghui Wu, Mike Schuster, Zhifeng Chen, Quoc~V Le, Mohammad Norouzi, Wolfgang
  Macherey, Maxim Krikun, Yuan Cao, Qin Gao, Klaus Macherey, et~al.
\newblock Google's neural machine translation system: Bridging the gap between
  human and machine translation.
\newblock {\em arXiv}, 2016.

\bibitem{wu2019adaframe}
Zuxuan Wu, Caiming Xiong, Chih-Yao Ma, Richard Socher, and Larry~S Davis.
\newblock Adaframe: Adaptive frame selection for fast video recognition.
\newblock In {\em CVPR}, 2019.

\bibitem{Xie_2018_ECCV}
Saining Xie, Chen Sun, Jonathan Huang, Zhuowen Tu, and Kevin Murphy.
\newblock Rethinking spatiotemporal feature learning: Speed-accuracy trade-offs
  in video classification.
\newblock In {\em ECCV}, 2018.

\bibitem{xu2017video}
Dejing Xu, Zhou Zhao, Jun Xiao, Fei Wu, Hanwang Zhang, Xiangnan He, and Yueting
  Zhuang.
\newblock Video question answering via gradually refined attention over
  appearance and motion.
\newblock In {\em ACM MM}, 2017.

\bibitem{xu2016msr}
Jun Xu, Tao Mei, Ting Yao, and Yong Rui.
\newblock Msr-vtt: A large video description dataset for bridging video and
  language.
\newblock In {\em CVPR}, 2016.

\bibitem{yang2019xlnet}
Zhilin Yang, Zihang Dai, Yiming Yang, Jaime Carbonell, Russ~R Salakhutdinov,
  and Quoc~V Le.
\newblock Xlnet: Generalized autoregressive pretraining for language
  understanding.
\newblock In {\em NeurIPS}, 2019.

\bibitem{yu2018joint}
Youngjae Yu, Jongseok Kim, and Gunhee Kim.
\newblock A joint sequence fusion model for video question answering and
  retrieval.
\newblock In {\em ECCV}, 2018.

\bibitem{yu2016video}
Youngjae Yu, Hyungjin Ko, Jongwook Choi, and Gunhee Kim.
\newblock Video captioning and retrieval models with semantic attention.
\newblock {\em arXiv}, 2016.

\bibitem{yu2017end}
Youngjae Yu, Hyungjin Ko, Jongwook Choi, and Gunhee Kim.
\newblock End-to-end concept word detection for video captioning, retrieval,
  and question answering.
\newblock In {\em CVPR}, 2017.

\bibitem{zhang2018cross}
Bowen Zhang, Hexiang Hu, and Fei Sha.
\newblock Cross-modal and hierarchical modeling of video and text.
\newblock In {\em ECCV}, 2018.

\bibitem{zhou2020unified}
Luowei Zhou, Hamid Palangi, Lei Zhang, Houdong Hu, Jason~J Corso, and Jianfeng
  Gao.
\newblock Unified vision-language pre-training for image captioning and vqa.
\newblock In {\em AAAI}, 2020.

\bibitem{zhou2017towards}
Luowei Zhou, Chenliang Xu, and Jason~J Corso.
\newblock Towards automatic learning of procedures from web instructional
  videos.
\newblock In {\em AAAI}, 2018.

\bibitem{zhu2020actbert}
Linchao Zhu and Yi Yang.
\newblock Actbert: Learning global-local video-text representations.
\newblock In {\em CVPR}, 2020.

\bibitem{zhu2015aligning}
Yukun Zhu, Ryan Kiros, Rich Zemel, Ruslan Salakhutdinov, Raquel Urtasun,
  Antonio Torralba, and Sanja Fidler.
\newblock Aligning books and movies: Towards story-like visual explanations by
  watching movies and reading books.
\newblock In {\em ICCV}, 2015.

\end{thebibliography}
}

\end{document}